\documentclass{bioinfo}
\copyrightyear{2017} \pubyear{2017}

\access{}
\appnotes{}

\usepackage{amsfonts}
\usepackage{caption}
\usepackage{subcaption}
\usepackage{multirow}
\usepackage[table,xcdraw]{xcolor}
\usepackage{booktabs}
\usepackage{tabularx}
\usepackage{listings}
\usepackage[margin=0.9in]{geometry}
\usepackage[hyphens]{url}
\usepackage{xspace}
\usepackage{url}
\usepackage[colorlinks,citecolor=blue]{hyperref}
\usepackage{color}
\usepackage{xcolor}
\usepackage{algorithm}
\usepackage{algorithmic}
\usepackage{graphicx}
\usepackage{amsmath} 
\usepackage{amssymb}

\newcommand{\hide}[1]{}
\newcommand{\xhdr}[1]{\vspace{1.7mm}\noindent{{\bf #1.}}}

\newcommand{\ohmnet}{{\em{OhmNet}}\xspace}

\makeatletter
\newcommand\fs@norules{\def\@fs@cfont{\bfseries}\let\@fs@capt\floatc@ruled
  \def\@fs@pre{}%
  \def\@fs@post{}%
  \def\@fs@mid{\kern3pt}%
  \let\@fs@iftopcapt\iftrue}
\makeatother 
\floatstyle{norules}
\restylefloat{algorithm}

\newcommand{\eg}{\emph{e.g.}\xspace}  
\newcommand{\ie}{\emph{i.e.}\xspace}

\graphicspath{{./FIG/}}

\begin{document}
\firstpage{1}

%\subtitle{Subject Section}
\subtitle{}

\title[Predicting multicellular function through multi-layer tissue networks]{Predicting multicellular function through \\ multi-layer tissue networks}
\author[Zitnik \& Leskovec]{Marinka Zitnik and Jure Leskovec\,$^{*}$}
\address{Department of Computer Science, Stanford University, Stanford, 94305, USA}

\corresp{$^\ast$To whom correspondence should be addressed.}

\history{}
\editor{}

\abstract{
\textbf{Motivation:} 
Understanding functions of proteins in specific human tissues is essential for insights into disease diagnostics and therapeutics, yet prediction of tissue-specific cellular function remains a critical challenge for biomedicine.\\[1mm]
\textbf{Results:} 
Here we present \ohmnet, a 
hierarchy-aware unsupervised node feature learning approach for multi-layer networks. We build a multi-layer network, where each layer represents molecular interactions in a different human tissue. \ohmnet then automatically learns a mapping of proteins, represented as nodes, to a neural embedding based low-dimensional space of features. \ohmnet encourages sharing of similar features among proteins with similar network neighborhoods and among proteins activated in similar tissues. 
The algorithm generalizes prior work, which generally ignores relationships between tissues, by modeling tissue organization with a rich multiscale tissue hierarchy.
We use \ohmnet to study multicellular function in a multi-layer protein interaction network of 107 human tissues. In 48 tissues with known tissue-specific cellular functions, \ohmnet provides more accurate predictions of cellular function than alternative approaches, and also generates more accurate hypotheses about tissue-specific protein actions. We show that taking into account the tissue hierarchy leads to improved predictive power. Remarkably, we also demonstrate that it is possible to leverage the tissue hierarchy in order to effectively transfer cellular functions to a functionally uncharacterized tissue. 
Overall, \ohmnet moves from flat networks to multiscale models able to predict a range of  phenotypes spanning cellular subsystems.
\\[2mm]
\textbf{Availability:}
Source code and datasets are available at \url{http://snap.stanford.edu/ohmnet}.\\[1mm]
\textbf{Contact:} jure@cs.stanford.edu
%
%\textbf{Supplementary information:} Supplementary data are available at \textit{Bioinformatics} online.
}

\maketitle

\section{Introduction}
\label{sec:intro}

A unified view of human diseases and cellular functions across a broad range of human tissues is essential, not only for understanding basic biology but also for interpreting genetic variation and developing therapeutic strategies~\citep{Yeger2015,Okabe2014,Greene2015,Gtex2015}.
In particular, the precise functions of proteins frequently depend on the tissue, and different proteins can have different cellular functions in different tissues~\citep{Lois2002,Rakyan2008,Magger2012,Guan2012,Fagerberg2014,Yeger2015,Hu2016}.

While our view of the human protein-protein interaction network as a key source for studying protein function, is constantly expanding, much less is known about networks that form in biologically important environments such as within distinct tissues or in specific diseases~\citep{Yeger2015}. Although incredibly influential, current computational methods for extracting functional information from protein interaction networks lack tissue specificity as they assume that cellular function is constant across organs and tissues~\citep{Barutcuoglu2006,Mostafavi2008,Radivojac2013,Stojanova2013,Kramer2014,Zitnik2015}. In other words, cellular functions in heart are assumed to be the same as functions in skin. The methods are, hence, less successful in constructing {\em accurate maps of both where and how proteins act.}  In particular, existing network-based methods are probably not the ultimate representation of human tissues for three reasons. (1) First, current methods for cellular function prediction on networks~\citep{Mostafavi2009,Radivojac2013,Zitnik2015,Vidulin2016} do not model networks with regards to patterns that span tissues, organs, and cellular systems. This means that a complex tissue involving a multiscale hierarchy of cellular subsystems is not readily captured by current models~\citep{Dutkowski2012,Carvunis2014}. (2) Second, many genome-scale functional maps~\citep{Lopes2011,Rolland2014,Kotlyar2015,Kitsak2016,Costanzo2016,Wang2016} are descriptive maps of physical or functional protein connectivity that do not, by themselves, predict cellular function. (3) Third, only few computational approaches~\citep{Magger2012,Guan2012,Ganegoda2014,Antanaviciute2015} used tissue-specific information to identify novel genes and relationships between genes. However, their focus was to leverage tissue specificity to improve prediction of global cellular functions and global gene-disease associations. As such, these approaches account for tissue specificity, but they do not resolve the challenge of predicting gene-function relationships that might be specific to a particular tissue. To be able to predict a range of tissue-specific functions one needs to design scalable multiscale models that can relate tissues to each other, extract rich feature representations for proteins in each tissue-specific network, and then use the extracted features for tissue-specific cellular function prediction.

\begin{figure}
 \begin{center}
\includegraphics[width=0.47\textwidth]{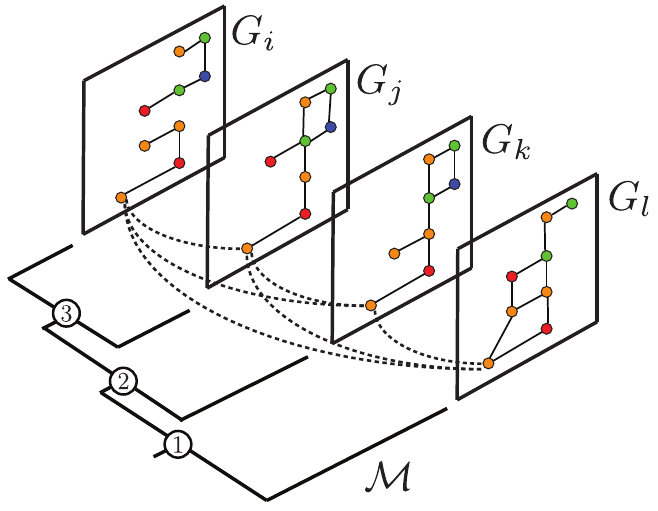}
 \caption{\textbf{A multi-layer network with four layers, where each layer represents a tissue-specific protein-protein interaction network.} The hierarchy $M$ encodes biological similarities between the tissues at multiple scales. \ohmnet embeds each node in a $d$-dimensional feature space, which we use for tissue-specific protein function prediction. For example, layers $G_i$, $G_j$, $G_k$ and $G_l$, might represent brain tissue-specific interaction networks in cerebrum, hypothalamus, tegmentum, and medulla. 
 }
 \label{fig:ohmnet}
 \end{center}
\end{figure}

\xhdr{Present work} 
We present \ohmnet, an algorithm for hierarchy-aware unsupervised feature learning in multi-layer networks. Our focus is on learning features of proteins in different tissues. We represent each tissue as a network, where nodes represent proteins. Tissue networks act as layers in a multi-layer network, where we use a hierarchy to model dependencies between the layers (\ie, tissues) (Figure~\ref{fig:ohmnet}). We then develop a computational framework that learns features of each node (\ie, protein) by taking into consideration connections between the nodes within each layer, together with inter-layer relationships between proteins active on different layers. More precisely, our approach embeds each protein in each tissue in a $d$-dimensional feature space such that proteins with similar network neighborhoods in similar tissues are embedded closely together. 

In \ohmnet, we define an objective function that is independent of the downstream prediction task, meaning that the feature representations are learned in a purely unsupervised way. This results in task-independent features, that, as we show, outperform task-specific approaches in predictive accuracy. Furthermore, since our features are not designed for a specific downstream prediction task, they generalize across a wide variety of tasks and tissues. For example, we use the learned features to study protein functions across different cellular systems (\eg, cell types, tissues, organs, and organ systems). 

\ohmnet builds on recent success of unsupervised representation learning methods based on neural architectures~\citep{Mikolov2013,Grover2016}. In particular, we develop a new form of structured regularization, which makes \ohmnet especially suitable for multi-layer interdependent networks. Our key contribution lies in modeling the tissue taxonomy constraints by encoding relationships between the tissues in a tissue hierarchy and then using the structured regularization with the tissue hierarchy (Figure~\ref{fig:ohmnet}). This way \ohmnet effectively learns multiscale feature representations for proteins that are consistent with the tissue hierarchy. 

Our experiments focus on three tasks defined on a multi-layer tissue network: (i) a multi-label node classification task, where every protein is assigned zero, one or more tissue-specific cellular functions; (ii) a transfer learning task, where we predict cellular functions for a protein in one tissue based on classifiers trained on features from other tissues; and (iii) a network embedding visualization task, where we create meaningful tissue-specific visualizations that lay out proteins on a two-dimensional space. Since the multiscale protein feature vectors returned by \ohmnet are task-independent, we use \ohmnet one time only to learn the features for proteins in every tissue and at every scale of the tissue hierarchy. We can then solve the cellular function prediction task for any tissue using the appropriate tissue-specific protein features.

We contrast \ohmnet's performance with that of state-of-the-art approaches for feature learning~\citep{Nickel2011,Tang2015,Cannistraci2013,Grover2016}, approaches for tissue-independent cellular function prediction~\citep{Mostafavi2008,Zuberi2013}, and approaches for prioritization of disease-causing genes in tissue-specific protein interaction networks~\citep{Magger2012,Guan2012}, which we adapted for the cellular function prediction task. We experiment with a multi-layer network having 107 genome-wide tissue-specific protein interaction layers, and we consider a tissue hierarchy describing 219 cellular systems in the human body. Experiments demonstrate that tissue-specific protein interaction layers provide the necessary protein and tissue context for predicting cellular function. \ohmnet outperforms alternative approaches by up to 14.9\% on multi-label classification and up to 20.3\% on transfer learning. Another notable finding is that \ohmnet outperforms alternative approaches, which are based on non-hierarchical versions of the same dataset, alluding to the benefits of modeling hierarchical tissue organization. We observe that neglecting the existence of tissues or aggregating tissue-specific interaction networks into a single network discards important biological information and affects performance on multi-label classification and transfer learning tasks. Finally, we exemplify the utility of \ohmnet for exploring the multiscale structure of tissues. In a case study on nine brain tissue networks, we show that \ohmnet's features inherently encode a multiscale brain organization.

The rest of the paper is organized as follows. In Section~\ref{sec:related}, we briefly survey related work in feature learning for networks. We present the technical details of \ohmnet in Section~\ref{sec:ohmnet}. In Section~\ref{sec:data}, we describe the multi-layer tissue network and the tissue hierarchy. We empirically evaluate \ohmnet in Section~\ref{sec:results} and conclude with directions for future work in Section~\ref{sec:conclusion}.

\section{Related work}
\label{sec:related}

We have seen in Section~\ref{sec:intro} that despite the abundance of methods for cellular function prediction, only a few, if any, take into account biologically important contexts given by human tissues. We now turn our focus to the problem of feature learning in networks.

Most approaches for automatic (\ie, non-hand-engineered) feature learning in networks can be categorized into matrix factorization and neural network embedding based approaches. In matrix factorization, a network is expressed as a data matrix where the entries represent relationships. The data matrix is projected to a low dimensional space using linear techniques based on SVD~\citep{Tang2012}, or non-linear techniques based on multi-dimensional scaling~\citep{Tenenbaum2000,Belkin2001,Hou2014}. These methods have two important drawbacks. First, they do not account for important structures typically exhibited in networks such as high sparsity and skewed degree distribution. Second, matrix factorization methods perform a global factorization of the data matrix while a local-centric method might often yield more useful feature representations~\citep{Kramer2014}.

Limitations of matrix factorization are overcome by neural network embeddings. Recent studies focused on embedding nodes into low-dimensional vector spaces by first using random walks to construct the network neighborhood of every node in the graph, and then optimizing an objective function with network neighborhoods as input~\citep{Perozzi2014,Tang2015,Grover2016}. The objective function is carefully designed to preserve both the local and global network structures. A state-of-the-art neural network embedding algorithm is the Node2vec algorithm~\citep{Grover2016}, which learns feature representations as follows: it scans over the nodes in a network, and for every node it aims to embed it such that the node's features can predict nearby nodes, that is, node's feature predict which other nodes are part of its network neighborhood. Node2vec can explore different network neighborhoods to embed nodes based on the principles of homophily (\ie, network communities) as well as structural equivalence (\ie, structural roles of nodes).

However, a challenging problem for neural network embedding-based methods is to learn features in multi-layer networks. Existing methods can learn features in multi-layer networks either by treating each layer independently of other layers, or by aggregating the layers into a single (weighted) network. However, neglecting the existence of multiple layers or aggregating the layers into a single network, alters topological properties of the system as well as the importance of individual nodes with respect to the entire network structure~\citep{De2016}. This is a major shortcoming of prior work that can lead to a wrong identification of the most versatile nodes~\citep{De2015} and overestimation of the importance of more marginal nodes~\citep{De2014}. As we shall show, this shortcoming also affects predictive accuracy of the learned features. Our approach \ohmnet overcomes this limitation since it learns features in a multi-layer network in the context of the entire system structure, bridging together different layers and generalizing methods developed for learning features in single-layer networks.

In biological domains, measures based on similarities of nodes' extended network neighborhoods are well established for predicting protein functions. Several approaches use graphlets~\citep{Przulj2007} to systematically describe network structure around each node. This is done by counting how many instances of small subgraph patterns occur in the network neighborhood of a given node. Graphlet-based methods, such as graphlet degree vectors~\citep{Hayes2013}, can thus be seen as an alternative approach for extracting feature representations for nodes. In contrast to neural embedding-based methods, such as \ohmnet, which learn continuous feature representations, graphlet-based methods return discrete counts of motif occurrences. Further, graphlet-based methods in their current form cannot be applied to multi-layer networks without collapsing the network layers into one network. 

Finally, there exists recent work for task-dependent feature learning based on graph-specific deep network architectures~\citep{Zhai2015,Li2015,Xiaoyi2014,Wang2016b}. Our approach differs from those approaches in two important ways. First, those architectures are task-dependent, meaning they directly optimize the objective function for a downstream prediction task, such as cellular function prediction in a particular tissue, using several layers of non-linear transformations. Second, those architectures do not model rich graph structures, such as multi-layer networks with hierarchies.

\begin{methods}
\section{Feature learning in multi-layer networks}
\label{sec:ohmnet}

We formulate feature learning in multi-layer networks as a maximum likelihood optimization problem. Let $V$ be a given set of $N$ nodes (\eg, proteins) $\{u_1, u_2, \dots, u_N\},$ and let there be $K$ types of edges (\eg, protein interactions in different tissues) between pairs of nodes $u_1, u_2, \dots, u_N$. A multi-layer network is a general system in which each biological context is represented by a distinct {\em layer} $i$ (where $i = 1, 2, \dots, K$) of a system (Figure~\ref{fig:ohmnet}). We use the term {\em single-layer network (layer)} for the network $G_i = (V_i, E_i)$ that indicates the edges $E_i$ between nodes $V_i \subseteq V$ within the same layer $i$. Our analysis is general and applies to any (un)directed, (un)weighted multi-layer network. 

We take into account the possibility that a node $u_k$ from layer $i$ can be related to any other node $u_h$ in any other layer $j$. We encode information about the dependencies between layers in a hierarchical manner that we use in the learning process. Let the hierarchy be a directed tree $\mathcal{M}$ defined over a set $M$ of elements by the parent-child relationships given by $\pi : M \rightarrow M,$ where $\pi(i)$ is the parent of element $i$ in the hierarchy (Figure~\ref{fig:ohmnet}). Let $T \subset M$ be the set of all leaves in the hierarchy. Let $T_i$ be the set of all leaves in the sub-hierarchy rooted at $i$. We assume that each layer $G_i$ is attached to one leaf in the hierarchy. As a result, the hierarchy $\mathcal{M}$ has exactly $K$ leaves. For convenience, let $C_i$ denote the set of all children of element $i$ in the hierarchy.

The problem of feature learning in a multi-layer network is to learn functions $f_1, f_2, \dots, f_K$, such that each function $f_i : V_i \rightarrow \mathbb{R}^d$ maps nodes in $V_i$ to feature representations in $\mathbb{R}^d$. Here, $d$ is a parameter specifying the number of dimensions in the feature representation of one node. Equivalently, $f_i$ is a matrix of $|V_i| \times d$ parameters. 

We proceed by describing \ohmnet, our approach for feature learning in multi-layer networks. \ohmnet has two components:
\begin{itemize}
\item {\bf single-layer network objectives}, in which nodes with similar network neighborhoods in each layer are embedded close together,
\item {\bf hierarchical dependency objectives}, in which nodes in nearby layers in the hierarchy are encouraged to share similar features.
\end{itemize}
We start by describing the model that considers the layers independently of each other. We then extend the model to encourage nodes which are nearby in the hierarchy to have similar features. 

\subsection{Single-layer network objectives}

We start by formalizing the intuition that nodes with similar network neighborhoods in each layer should share similar features. For that, we specify one objective for each layer in a given multi-layer network. We shall later discuss how \ohmnet incorporates the dependencies between different layers. 

Our goal is to take layer $G_i$ and learn $f_i$ which embeds nodes from similar network regions, or nodes with similar structural roles, closely together. In \ohmnet, we aim to achieve this goal by specifying the following objective function for each layer $G_i$. Given a node $u \in V_i$, the objective function $\omega_i$ seeks to predict, which nodes are members of $u$'s network neighborhood $N_{i}(u)$ based on the learned node features $f_i$:
\begin{equation}
\omega_i(u) = \log Pr(N_{i}(u) | f_i(u)),
\end{equation}
where the conditional likelihood of every node-neighborhood node pair is modeled independently as:
\begin{equation}
Pr(N_{i}(u) | f_i(u)) = \prod_{v \in N_i(u)} Pr(v | f_i(u)).
\end{equation}
The conditional likelihood is a softmax unit parameterized by a dot product of nodes' features, which corresponds to a single-layer feed-forward neural network: 
\begin{equation}
Pr(v | f_i(u)) = \frac{\exp(f_i(v) f_i(u))}{\sum_{z \in V_i} \exp(f_i(z) f_i(u))}. 
\end{equation}
Given a node $u$, maximization of $\omega_i(u)$ tries to maximize classification of nodes in $u$'s network neighborhood based on $u$'s learned representation. 

The objective $\Omega_i$ is defined for each layer $i$:
\begin{equation}
\Omega_i = \sum_{u \in V_i} \omega_i(u), \;\;\; \textrm{for } i=1, 2, \dots, K.
\label{eq:omega-def}
\end{equation}
The objective is inspired by the intuition that nodes with similar network neighborhoods tend to have similar meanings, or roles, in a network. It formalizes this intuition by encouraging nodes in similar network neighborhoods to share similar features.

We found that a flexible notion of a network neighborhood $N_i$ is crucial to achieve excellent predictive accuracy on a downstream cellular function prediction task~\citep{Grover2016}. For that reason, we use a randomized procedure to sample many different neighborhoods of a given node $u$. Technically, the network neighborhood $N_{i}(u)$ is a set of nodes that appear in an appropriately biased random walk defined on layer $G_i$ and started at node $u$~\citep{Grover2016}. The neighborhoods $N_i(u)$ are not restricted to just immediate neighbors but can have vastly different structures depending on the sampling strategy. 

Next, we expand \ohmnet's single-layer network objectives to leverage information provided by the tissue taxonomy and this way inform embeddings across different layers.

\subsection{Hierarchical dependency objectives}

So far, we specified $K$ layer-by-layer objectives each of which estimates node features in its layer independently of node features in other layers. This means that nodes in different layers representing the same entity have features that are learned independently of each other. 

To harness the dependencies between the layers, we expand \ohmnet with terms that encourage sharing of protein features between the layers. Our approach is based on the assumption that nearby layers in the hierarchy are semantically close to each other and hence proteins/nodes in them should share similar features. For example, in the tissue multi-layer network, we model the fact that the ``medulla'' layer is part of the ``brainstem'' layer, which is, in turn, part of the ``brain'' layer.  We use the dependencies among the layers to define a joint objective for regularization of the learned features of proteins. 

We propose to use the hierarchy in the learning process by incorporating a recursive structure into the regularization term for every element in the hierarchy $\mathcal{M}$. Specifically, we propose the following form of regularization for node $u$ that resides in element $i$ of the hierarchy $\mathcal{M}$:
\begin{equation}
c_i(u) = \frac{1}{2} \| f_i(u) - f_{\pi(i)}(u) \|_2^2.
\end{equation}
This recursive form of regularization enforces the features of node $u$ in the hierarchy $i$ to be similar to the features of node $u$ in $i$'s parent $\pi(i)$ under the Euclidean norm. When regularizing features of all nodes across all elements of the hierarchy, we obtain:
\begin{equation}
C_i = \sum_{u \in L_i} c_i(u), \;\;\; \textrm{where } L_i = \cup_{j \in T_i} V_j
\end{equation}
In words, we specify the features for both leaf as well as internal, \ie, non-leaf, elements in the hierarchy, and we regularize the features of sibling (\ie, sharing the same parent) hierarchy elements towards features in the common parent element in the hierarchy. 

\xhdr{Node features at multiple scales}
It is important to notice that \ohmnet's structured regularization allows us to learn feature representations at multiple scales. For example, consider a multi-layer network in Figure~\ref{fig:hie-toy}, consisting of four layers that are interrelated by a two-level hierarchy. \ohmnet learns the mappings $f_i$, $f_j$, $f_k$, and $f_l$ that map nodes in each layer into a $d$-dimensional feature space. Additionally, \ohmnet also learns the mapping $f_2$ representing features for nodes appearing in the hierarchy leaves $T_2$, \ie, $V_i \cup V_j$, at an intermediate scale, and the mapping $f_1$ representing features for nodes appearing in the hierarchy leaves $T_1$, \ie, $V_i \cup V_j \cup V_k \cup V_l$, at the highest scale. 

\begin{figure}
 \begin{center}
\includegraphics[width=0.4\textwidth]{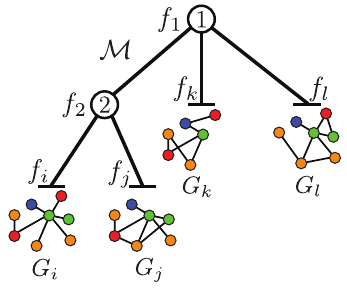}
 \caption{\textbf{A multi-layer network with four layers.} Relationships between the layers are encoded by a two-level hierarchy $\mathcal{M}$. Leaves of the hierarchy correspond to the network layers. Given networks $G_i$ and hierarchy $\mathcal{M}$, \ohmnet learns node embeddings captured by functions $f_i$.
} 
 \label{fig:hie-toy}
 \end{center}
\end{figure}

The modeling of relationships between layers in a multi-layer network has several implications:
\begin{itemize}
\item First, the model encourages nodes which are in nearby layers in the hierarchy to share similar features.
\item Second, the model shares statistical strength across the hierarchy as nodes in different layers representing the same protein share features through ancestors in the hierarchy.
\item Third, this model is more efficient than the fully pairwise model. In the fully pairwise model, the dependencies between layers are modeled by pairwise comparisons of nodes across all pairs of layers, which takes $O(K^2 N)$ time, where $K$ is the number of layers and $N$ is the number of nodes.  In contrast, \ohmnet models inter-layer dependencies according to the parent-child relationships specified by the hierarchy, which takes only $O(|M|N)$ time. Since \ohmnet's hierarchy is a tree, it holds that $|M| \ll K^2$, meaning that the proposed model scales more easily to large multi-layer networks than the fully pairwise model. 
\item Finally, the hierarchy is a natural way to represent and model biological systems spanning many different biological scales~\citep{Carvunis2014,Greene2015,Yu2016}. 
\end{itemize}

\subsection{Full \ohmnet model}

Given a multi-layer network consisting of layers $G_1, G_2, \dots, G_K$, and a hierarchy encoding relationships between the layers, the \ohmnet's goal is to learn the functions $f_1, f_2, \dots, f_K$ that map from nodes in each layer to feature representations. \ohmnet achieves this goal by fitting its feature learning model to a given multi-layer network and a given hierarchy, \ie, by finding the mapping functions $f_1, f_2, \dots, f_K$ that maximize the data likelihood. 

Given the data, \ohmnet aims to solve the following maximum likelihood optimization problem:
\begin{equation}
\max_{f_1, f_2, \dots, f_{|M|}} \sum_{i \in T} \Omega_i - \lambda \sum_{j \in M} C_j, 
\label{eq:opt-problem}
\end{equation}
which includes the single-layer network objectives for all network layers, and the hierarchical dependency objectives for all hierarchy elements. In Eq.~(\ref{eq:opt-problem}), parameter $\lambda$ is a user-specified parameter representing the regularization strength. While the optimization problem in Eq.~(\ref{eq:opt-problem}) is non-convex due to the non-convexity of the single-layer objective~\citep{Grover2016}, stochastic gradient with negative sampling can be used to efficiently solve the problem.  

One appealing property of \ohmnet is that by solving the problem in Eq.~(\ref{eq:opt-problem}) we obtain estimates for functions $f_1, f_2, \dots, f_K$ located in the leaf elements of the hierarchy (\ie, layers of a given multi-layer network), as well as estimates for functions $f_{K+1}, f_{K+2}, \dots, f_{|M|}$ located in the internal elements of the hierarchy.

\subsection{The \ohmnet algorithm}

The pseudocode for \ohmnet is given in Algorithm~\ref{alg:ohmnet}. 

In the first phase, \ohmnet applies the Node2vec's algorithm~\citep{Grover2016} to construct network neighborhoods for each node in every layer. Given a layer $G_i$ and a node $u \in V_i$, the algorithm simulates a user-defined number of fixed length random walks started at node $u$ (step 4 in Algorithm~\ref{alg:ohmnet}). 

In the second phase, \ohmnet uses an iterative approach in which features associated with each object in the hierarchy are iteratively updated by fixing the rest of the features. The iterative approach has the advantage that it can easily incorporate the closed-form updates developed for the internal objects of the hierarchy (step 11 in Algorithm~\ref{alg:ohmnet}), thereby accelerating the convergence of \ohmnet algorithm. For each leaf object $i$, \ohmnet isolates the terms in the optimization problem in Eq.~(\ref{eq:opt-problem}) that depend on the model parameters defining function $f_i$. \ohmnet then optimizes Eq.~(\ref{eq:opt-problem}) by performing one epoch of stochastic gradient descent (SGD1) over $f_i$'s model parameters (step 15 in Algorithm~\ref{alg:ohmnet}).

The two phases of \ohmnet are executed sequentially. The \ohmnet algorithm scales to large multi-layer networks because each phase is parallelizable and executed asynchronously. The choice to model the dependencies between network layers using the hierarchical model requires $O(|M| N)$ time instead of the fully pairwise model, which requires $O(K^2 N)$ time.

\begin{algorithm}
 \caption{\textbf{The \ohmnet algorithm.}}
 \begin{algorithmic}[1]
 \renewcommand{\algorithmicrequire}{\textbf{Input:}}
 \renewcommand{\algorithmicensure}{\textbf{Output:}}
 \REQUIRE Multi-layer network, $(G_1, G_2, \dots, G_K)$ with $G_i = (V_i, E_i)$, Hierarchy, $\mathcal{M}$, Feature representation size, $d$, Network neighborhood strategy, $S$, Regularization strength, $\lambda$\\[1mm]
 \FOR {$i \in T$} 
 \FOR {$u \in V_i$}
 \STATE $N_{i}(u) = \textrm{Node2vecWalk}(G_i, u, S)$ \citep{Grover2016}
 \ENDFOR
 \ENDFOR
  \WHILE {$f_1, f_2, \dots, f_{|M|}$ not converged}
  \FOR {$i \in M$}
  \IF {$i \in T$}
  \FOR {$u \in V_i$}
  \STATE $f_i(u) \!= \!\textrm{SGD1}(N_{i}(u), d, \lambda)$ by Eq.~(\ref{eq:opt-problem})
  \ENDFOR
  \ELSE
  \FOR {$u \in \cup_{j \in T_i} V_j$}
  \STATE $f_i(u) \!=\! \frac{1}{|C_{i}| + 1} ( f_{\pi(i)}(u) \!+\! \sum_{c \in C_{i}} \! f_c(u))$
  \ENDFOR
  \ENDIF
  \ENDFOR
  \ENDWHILE
 \RETURN $f_1, f_2, \dots, f_{|M|}$
 \end{algorithmic}
 \label{alg:ohmnet}
 \end{algorithm}

\end{methods}

\section{Tissue-specific interactome data}
\label{sec:data}

To construct the human protein-protein interaction (PPI) network, tissue-specific network layers, tissue hierarchy, and tissue-specific gene-function relationships, we downloaded and used standard protein, tissue, and function information from various reputable data sources. 

\subsection{Tissue hierarchy}\label{sec:tissue-hierarchy}

\begin{figure}  
 \begin{center} 
\includegraphics[width=\linewidth]{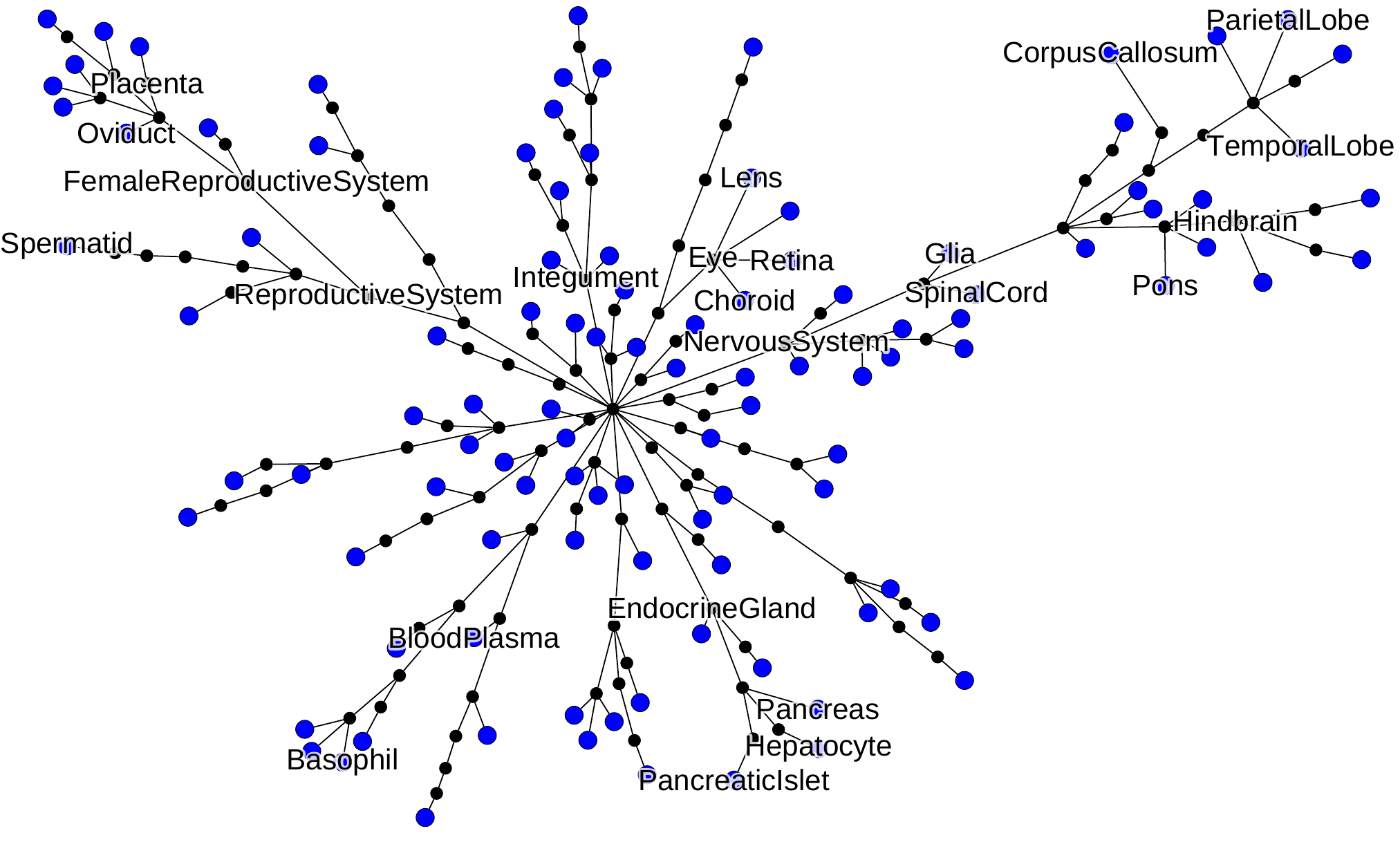}
 \caption{\textbf{The tissue hierarchy considered in this work.} The tissue hierarchy is a directed tree defined over $|M| = 219$ tissue terms from the BRENDA Tissue Ontology. Edges in the tree point from children to parents based on ontological relationships: ``develops\_from'', ``is\_a'', ``part\_of'', and ``related\_to''. The $K = 107$ tissues with tissue-specific protein interaction networks are the blue leaves in the tree.}
 \label{fig:tissue-hierarchy}
 \end{center}
\end{figure}

We retrieved the mapping of tissues in the Human Protein Reference Database (HPRD)~\citep{Prasad2009} to tissues in the BRENDA Tissue Ontology~\citep{Chang2014} from \cite{Greene2015}. The data is provided as a supplementary dataset in \cite{Greene2015}. The hierarchical relationships between tissues were then determined by the directed acyclic graph structure of the BRENDA Tissue Ontology. Examples of tissues included: muscle, adrenal cortex, bone marrow, and spleen (Figure~\ref{fig:tissue-hierarchy}).

\subsection{Tissue-specific interaction networks}\label{sec:interactome-data}

We took the gene-to-tissue mapping compiled by \cite{Greene2015}. \citeauthor{Greene2015} mapped genes to HPRD tissues based on low-throughput tissue-specific gene expression data. The gene-to-tissue mapping was then combined with the human PPI network. The resulting multi-layer tissue network had 107 layers, each layer corresponded to a PPI network specific to a particular tissue. Details are provided next. 

The human PPI network was collected from \cite{Orchard2013,Rolland2014,Chatr2015,Prasad2009,Ruepp2010,Menche2015}. Considered were physical protein-protein interactions with supported by experimental evidence. It should be noted that interactions based on gene expression and evolutionary data were not considered. The global (unweighted) human PPI network has 21,557 proteins interconnected by 342,353 interactions. The reader is referred to \cite{Menche2015} for a detailed description of the data. 

For each of 107 tissues, a tissue-specific human PPI network was constructed based on the global PPI network. For a given tissue, every edge in the global PPI network was labeled as specifically co-expressed in that tissue using the criterion developed by \cite{Greene2015}. \citeauthor{Greene2015} labeled each edge as specifically co-expressed if either both proteins are specific to that tissue or one protein is tissue-specific and the other is ubiquitous. Lists of specifically co-expressed proteins were retrieved from \cite{Greene2015}. Finally, the PPI network specific to a particular tissue is a subnetwork of the global PPI network, induced by the set of specifically co-expressed edges in that tissue.

\subsection{Tissue-specific cellular functions and gene annotations}
\label{sec:go-data}

Associations between tissues and cellular functions were retrieved from \cite{Greene2015}. \citeauthor{Greene2015} manually curated biological processes in the Gene Ontology\citep{Ashburner2000} (GO) and mapped them to tissues in the BRENDA Tissue Ontology~\citep{Chang2014} based on whether a given biological process is specifically active in a given tissue. The data is provided as a supplementar dataset in \cite{Greene2015}. An example of a cellular function-tissue pair is ``low-density lipoprotein particle remodeling'' in the blood plasma tissue. 

All gene annotations were propagated along the ontology hierarchy. Considered are functions with at least 15 annotated proteins~\citep{Guan2012}. In total, there are 584 tissue-specific cellular functions covering 48 distinct tissues. Each tissue-specific function is assigned to one or more leaves in the tissue hierarchy (Section~\ref{sec:tissue-hierarchy}). 

\section{Results}
\label{sec:results}

The \ohmnet's objective in Eq.~(\ref{eq:opt-problem}) is independent of any downstream task. This flexibility offered by \ohmnet makes the learned feature representations suitable for a variety of analytics tasks discussed below.

\subsection{Prediction of tissue-specific cellular functions}\label{sec:classify}

\xhdr{Experimental setup}
We view the problem of predicting cellular functions as solving a multi-label node classification task. Here, every node (\ie, protein) is assigned one or more labels (\ie, cellular functions from the GO) from a finite set of labels (\ie, all cellular functions in the GO, see Section~\ref{sec:go-data}). 

We apply \ohmnet, which for every node in every layer learns a separate feature vector in an unsupervised way. Thus, for every layer and every function we then train a separate one-vs-all linear classifier using the modified Huber loss with elastic net regularization. Using cross validation, we observe 90\% of proteins and all their cellular functions across the layers during the training phase. The task is then to predict the tissue-specific functions for the remaining 10\% of proteins. 

We evaluate the performance of \ohmnet against the following feature learning approaches:
\begin{itemize}
\item RESCAL tensor decomposition~\citep{Nickel2011}: This is a tensor factorization approach that takes the multi-layer network structure into account. Given $X_i$, a normalized Laplacian matrix of layer $G_i$, matrix $X_i$ is factorized as: $X_i = A R_i A^T$, for $i = 1, 2, \dots, K.$ Here, matrix $A$ contains $d$-dimensional  feature representation for nodes.
\item Minimum curvilinear embedding~\citep{Cannistraci2013}: This is a non-linear unsupervised framework that embeds nodes in a low-dimensional space. The approach was originally developed for protein interaction prediction, aiming to embed protein pairs representing good candidate interactions closer to each other. It utilizes a network denoising method as well as structural information provided by the PPI network topology.
\item LINE~\citep{Tang2015}: This approach first learns $d/2$ dimensions based on immediate network neighbors of nodes, and then the next $d/2$ dimensions based on network neighbors at a 2-hop distance. 
\item Node2vec~\citep{Grover2016}: This approach learns $d$-dimensional features for nodes based on a biased random walk procedure that flexibly explores network neighborhoods of nodes.
\end{itemize}
In addition, we evaluate the performance of \ohmnet against the following tissue-specific/agnostic function prediction approaches:
\begin{itemize}
\item GeneMania~\citep{Zuberi2013}: This is a supervised approach that takes a multi-layer network as input and directly predicts cellular functions in two separate phases. In the first phase, it aggregates the layers into one weighted network by weighting the layers according to their utility for predicting a given function. It then uses a label propagation algorithm on the weighted network to predict the function.
\item Tissue-specific network propagation~\citep{Magger2012}: This approach assigns a prior score to proteins associated with known functions that are phenotypically similar to the query function. This score is then propagated through a network in an iterative process. The approach was  developed for tissue-specific disease gene prioritization. 
\item Network-based tissue-specific SVM~\citep{Guan2012}: This approach adopts the network-based candidate gene prediction scheme. Essentially, the connection weights in a network to all positive examples (\ie, genes already known to be related to a phenotype) are utilized as features for linear support vector machine (SVM) classification. The approach was developed for tissue-specific phenotype and disease gene prioritization. 
\end{itemize}
The parameter settings for every approach are determined using internal cross validation procedure with a grid search over candidate parameter values. Specifically, $d=128$ is used in all experiments.

Last, we aim to evaluate the benefit of our proposed multi-layer representation of the tissue networks. To this end we also consider two additional network representations:

\begin{itemize}
\item Independent layers: This approach learns features for nodes in each layer by running LINE or Node2vec algorithm on one layer at a time and independently of other layers in the network.
\item Collapsed layers: This approach first aggregates the layers into a single network by connecting nodes representing the same entity in different layers to each other. It then learns feature for nodes in the aggregated network.
\end{itemize}

\begin{figure}  
 \begin{center} 
\includegraphics[width=\linewidth]{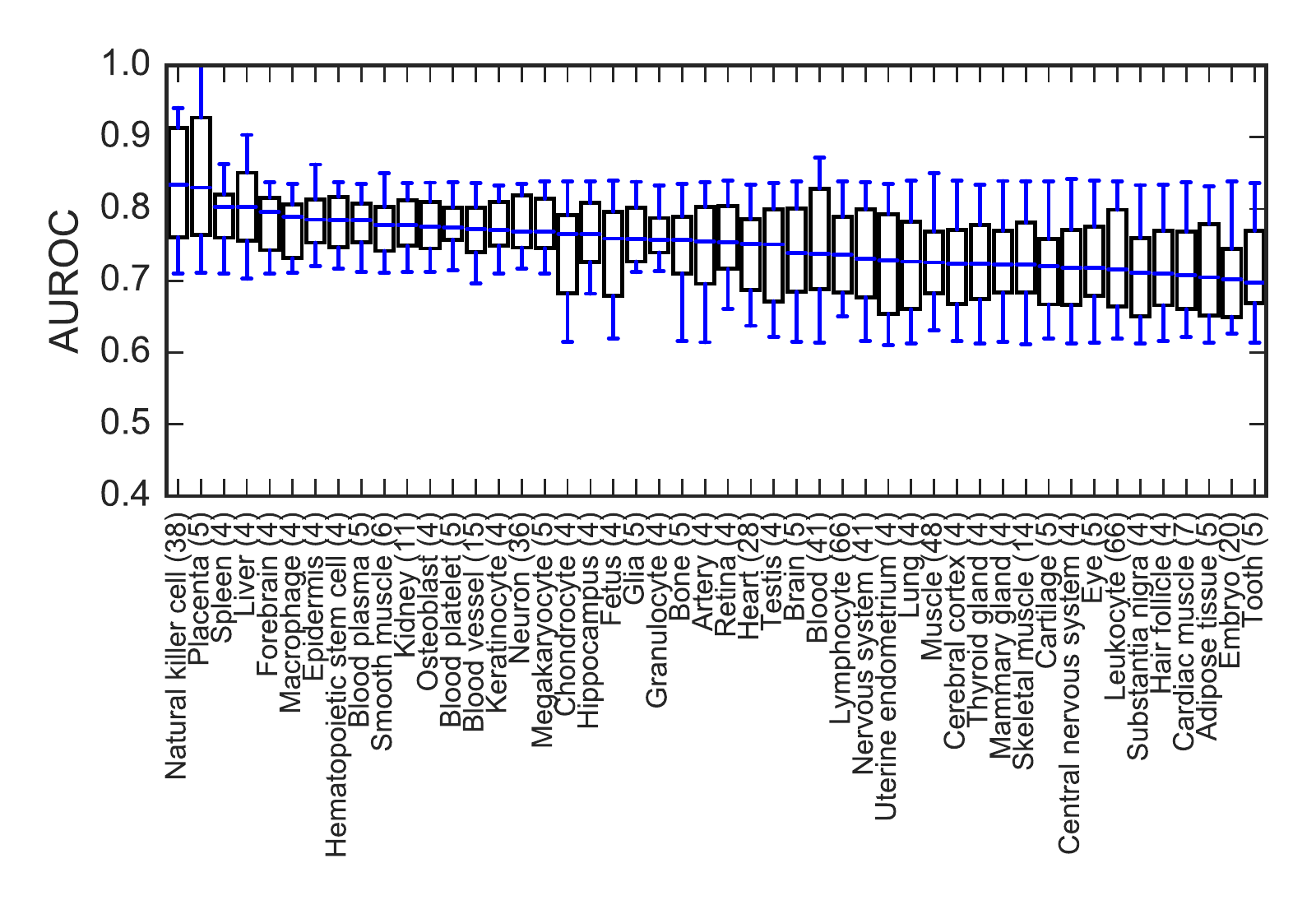}
 \caption{\textbf{Area under ROC curve (AUROC) scores for tissue-specific cellular function prediction by \ohmnet.} Numbers in the brackets are counts of tissue-specific cellular functions per tissue.}
 \label{fig:tissue-auroc}
 \end{center}
\end{figure}

\xhdr{Experimental results}
Table~\ref{tab:other-approaches} and Figure~\ref{fig:tissue-auroc} give the area under the curve (AUC) scores of tissue-specific protein function prediction.

From the results, we see how modeling the tissues and their hierarchy spanning multiple biological scales allows \ohmnet to outperform other benchmark approaches. \ohmnet outperforms GeneMania~\citep{Mostafavi2008,Zuberi2013} by 10.7\%, which can be explained by GeneMania's inability to weight layers in the tissue network according to a multiscale tissue organization that is consistent with the tissue taxonomy constraints. We also compared \ohmnet to two other methods~\citep{Guan2012,Magger2012} that were so far demonstrated as useful for mining tissue-specific protein relationships. \ohmnet has produced more accurate predictions, surpassing other methods by up to 12.0\% (AUROC) and up to 26.8\% (AUPRC). 

Independent modeling of the layers showed worse performance than collapsing the layers into one network. We observed that Collapsed LINE achieved a gain of 3.3\% over Independent LINE, and Collapsed Node2vec achieved a gain of 7.4\% over Independent Node2vec. However, approaches that neglect the existence of tissues or collapse tissue-specific protein interaction networks into a single network discard important information about the rich hierarchy of biological systems, giving \ohmnet a 14.0\% gain over Collapsed LINE, and a 8.5\% gain over Collapsed Node2vec in AUC scores. This result is a good illustration of how tissue specificity is related to specialization of protein function~\citep{Greene2015}, and approaches able to directly profile proteins' distinct interaction neighborhoods in different tissues can leverage this specificity to generate more accurate hypotheses about tissue-specific protein actions. 

\begin{table}
\processtable{Area under ROC curve (AUROC) and area under precision-recall curve (AUPRC) scores for tissue-specific cellular function prediction. Values in the brackets are halves of the interquartile distance. \ohmnet's results are statistically significant with a p-value of less than 0.05.
\label{tab:other-approaches}}{
\setlength{\tabcolsep}{4pt}
\begin{tabular}{lcc}\toprule	
  Approach & AUROC & AUPRC \\\midrule
  Tensor decomposition & \multirow{2}{*}{0.674 ($\pm$ 0.124)} & \multirow{2}{*}{0.235 ($\pm$ 0.052)} \\
  \citep{Nickel2011} & & \\
  Minimum curvilinear embedding  & \multirow{2}{*}{0.674 ($\pm$ 0.064)} & \multirow{2}{*}{0.248 ($\pm$ 0.071)} \\
  \citep{Cannistraci2013} & & \\
  Independent LINE & \multirow{2}{*}{0.642 ($\pm$ 0.053)} & \multirow{2}{*}{0.261 ($\pm$ 0.068)} \\
  \citep{Tang2015} & & \\
  Collapsed LINE & \multirow{2}{*}{0.663 ($\pm$ 0.047)} & \multirow{2}{*}{0.271 ($\pm$ 0.053)} \\
  \citep{Tang2015} & & \\
  Independent Node2vec & \multirow{2}{*}{0.649 ($\pm$ 0.063)} & \multirow{2}{*}{0.283 ($\pm$ 0.052)} \\
  \citep{Grover2016} & & \\
  Collapsed Node2vec & \multirow{2}{*}{0.697 ($\pm$ 0.085)} & \multirow{2}{*}{0.298 ($\pm$ 0.061)} \\
  \citep{Grover2016} & & \\\midrule
  GeneMania & \multirow{2}{*}{0.683 ($\pm$ 0.077)} & \multirow{2}{*}{0.274 ($\pm$ 0.094)} \\
  \citep{Zuberi2013} & & \\
  Network-based tissue-specific SVM & \multirow{2}{*}{0.701 ($\pm$ 0.091)} & \multirow{2}{*}{0.281 ($\pm$ 0.059)} \\
  \citep{Guan2012} & & \\
  Tissue-specific network propagation & \multirow{2}{*}{0.675 ($\pm$ 0.051)} & \multirow{2}{*}{0.265 ($\pm$ 0.083)} \\
  \citep{Magger2012} & & \\\midrule
  \ohmnet & \multirow{2}{*}{0.756 ($\pm$ 0.067)} & \multirow{2}{*}{0.336 ($\pm$ 0.045)} \\
  (Section~\ref{sec:ohmnet}) \\\botrule
\end{tabular}}{}
\end{table}

\subsection{Transfer of cellular functions to a new tissue}\label{sec:transfer}

\xhdr{Experimental setup}
In the transfer learning setting, we attempt to transfer knowledge learned in one or more {\em source layers} and use it for prediction in a {\em target layer}. 	

As before, we apply \ohmnet to obtain a separate feature vector for every node and every layer in an unsupervised way. We then consider, in turn, every tissue as a target layer and {\em all other tissues} as source layers. For every function and every source layer, we train a separate classifier using the same classification model as in Section~\ref{sec:classify}. We then predict functions for the target layer using only classifiers trained on the source layers. That is, we aim to predict cellular functions taking place in the target tissue without having access to any cellular function gene annotation in that tissue, \ie, we pretend the target tissue has no annotations. Prediction for one node in the target layer is the weighted average of predictions of the classifiers trained on source layers. Weights reflect hierarchy-based distances of source tissues from the target tissue. They are determined by the closed-form expressions mathematically equivalent to \ohmnet's regularization (details omitted due to space constraints).  

\begin{table}
\processtable{Area under ROC curve (AUROC) scores for transfer learning. Shown are the scores for ten tissues with best performance on cellular function prediction task. ``Non-transfer'': a classifier is trained on a target tissue and then used to predict cellular functions in the same tissue (Section~\ref{sec:classify}). ``Transfer'': classifers are trained on all non-target tissues and then used to predict cellular functions in the target tissue (Section~\ref{sec:transfer}). \label{tab:transfer}}{
\setlength{\tabcolsep}{5pt}
\begin{tabular}{l|c|c}\toprule
  Target tissue & AUROC (Non-transfer) & AUROC (Transfer) \\\midrule
Natural killer cell & 0.834 ($\pm$ 0.076) & 0.776 ($\pm$ 0.063) \\
Placenta & 0.830 ($\pm$ 0.082) & 0.758 ($\pm$ 0.068) \\
Spleen & 0.803 ($\pm$ 0.030) & 0.779 ($\pm$ 0.043) \\
Liver & 0.803 ($\pm$ 0.047) & 0.741 ($\pm$ 0.025) \\
Forebrain & 0.796 ($\pm$ 0.036) & 0.755 ($\pm$ 0.037) \\
Macrophage  & 0.789 ($\pm$ 0.037) & 0.724 ($\pm$ 0.024) \\
Epidermis & 0.785 ($\pm$ 0.030) & 0.749 ($\pm$ 0.032) \\
Hematopoietic stem cell & 0.784 ($\pm$ 0.035) & 0.744 ($\pm$ 0.036) \\
Blood plasma & 0.784 ($\pm$ 0.027) & 0.703 ($\pm$ 0.039) \\
Smooth muscle & 0.778 ($\pm$ 0.031) & 0.729 ($\pm$ 0.041) \\\midrule
Average & 0.799 & 0.746 \\\botrule
\end{tabular}}{}
\end{table}

\xhdr{Experimental results}
Table~\ref{tab:transfer} shows the classification accuracy results for transfer learning based on \ohmnet. Since transfer tasks are more difficult than non-transfer tasks (Section~\ref{sec:classify}), it is expected that the AUC scores will decrease on transfer tasks. Results in Table~\ref{tab:transfer} confirm these expectations; however, we observe a very graceful degradation in performance leading to an only 7\% average decrease in the AUC scores. We get the smallest performance differences for target tissues with many biologically similar source tissues (\ie, source layers) in the tissue network. For example, performance difference for the forebrain is only 5.2\%, which is due to the fact that there are nine other layers in the tissue network closely related to the forebrain, such as the cerebellum and the midbrain. Considering all 48 tissues with tissue-specific cellular functions, \ohmnet outperforms all comparison methods on most transfer tasks, achieving a gain of up to 20.3\% over the closest benchmark in AUC scores (scores not shown). Notice that we exclude GeneMania in the comparison because it is not amenable to transfer learning. This result suggests that considering the relationships between tissues when learning features for proteins has a significant impact on transfer performance.

Generally speaking, we observed that the transferability of classifiers decreased when the tree-based distance between the source and the target tissue in the tissue hierarchy increased, which is consistent with the empirical evidence in transfer learning~\citep{Yosinski2014}. This also matches our intuition that a source tissue should be most informative for predicting cellular functions in an anatomically close target tissue (\eg, source and target tissues are both part of the same organ). 

\subsection{The multiscale model of brain tissues}\label{sec:case-study}

We have seen in Section~\ref{sec:tissue-hierarchy} that human tissues have a multi-level hierarchical organization. The tissue hierarchy categorizes tissues into: cell types, groups of cells with similar structure and function; organs, groups of tissues that work together to perform a specific activity; and organ systems, groups of two or more tissues that work together for the good of the entire body. We now aim to empirically demonstrate this fact and show that \ohmnet in fact can discover embeddings that obey this organization.

\begin{figure*}  
 \begin{center}
\includegraphics[width=\textwidth]{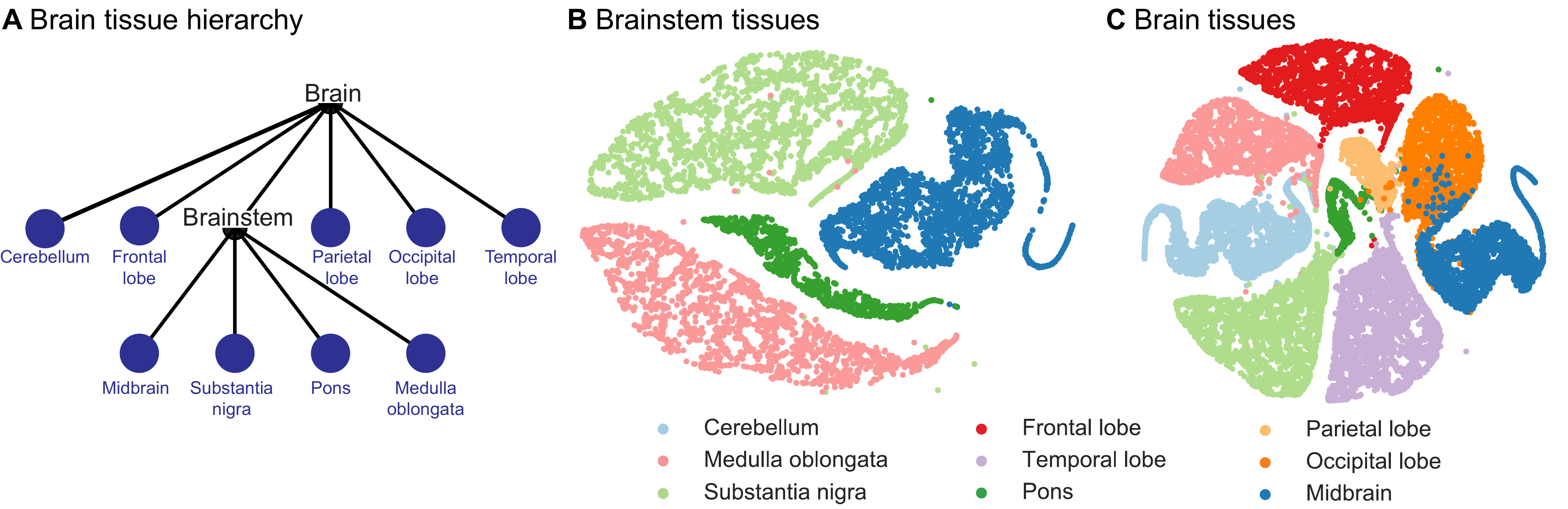}
 \caption{\textbf{Visualization of the brain tissue-specific protein interaction networks.} {\bf A.} The two-level brain tissue hierarchy as specified by the BRENDA Tissue Ontology~\citep{Chang2014} and used in the case study in Section~\ref{sec:case-study}. Leaves of the hierarchy (in blue) represent nine brain tissues each of which is associated with a tissue-specific protein interaction network. {\bf B.} Visualization of the brainstem-specific networks. The proteins are mapped to the 2-D space using the t-SNE package with learned features as input. Color of a node indicates the tissue of the protein.  {\bf C.} Visualization of the brain-specific networks. The proteins are mapped and colored using the same procedure as in B.}
 \label{fig:brain-case}
 \end{center}
\end{figure*}

We first construct a multi-layer brain network by integrating nine brain-specific protein interaction networks (\eg, the cerebellum, frontal lobe, brainstem, and other brain tissues). Each of nine brain-specific networks is one layer in the multi-layer network. The layers are organized according to a two-level hierarchy (Figure~\ref{fig:brain-case}{\color{red}a}). We run \ohmnet on this multi-layer network to find node features in a purely unsupervised way. We then map the nodes to the 2-D space based on the learned features. This way we assign every node in every layer to a point in the two-dimensional space based solely on the node's learned features. We then visualize the points and color them based on the layer they belong to. 

Figure~\ref{fig:brain-case}{\color{red}b} shows the example for the brainstem tissues: substantia nigra, pons, midbrain, and medulla oblongata. Laying out these tissue-specific networks is very challenging as the four brainstem tissues are very closely related to each other in the human body. However, the visualization using \ohmnet performs quite well. Notice how points of the same color are closely distributed, and how well regions of the same color are separated from each other. In the brainstem example, this means that \ohmnet generates a meaningful layout of the brainstem tissue-specific networks, in which proteins belonging to the same tissues are clustered together. 

Figure~\ref{fig:brain-case}{\color{red}c} shows the example for the brain, which is located one level up from the brainstem in the tissue hierarchy. Again, \ohmnet produces a meaningful layout of the nine brain tissue-specific networks. 

Additionally, we repeated this analysis by visualizing protein features learned by running principal component analysis (PCA) or non-negative matrix factorization (NMF) algorithm on the brain-specific PPI networks. Acknowledging the subjective nature of this analysis, we observed that visualizations using PCA or NMF were not very meaningful, as proteins belonging to the same tissue were not clustered together (data not shown). 

\ohmnet's result in Figure~\ref{fig:brain-case} is especially appealing because of two reasons. First, it shows that \ohmnet can learn node features that adhere to a given hierarchy of layers. In the brain example, \ohmnet learns the protein features that expose the multiscale tissue hierarchy. Second, it shows that \ohmnet can generate meaningful visualizations of network embeddings despite the fact that \ohmnet's objective is independent of the visualization task. 

\section{Conclusion}
\label{sec:conclusion}

We presented \ohmnet, an approach for unsupervised feature learning in multi-layer networks. We use \ohmnet to learn state-of-the-art task-independent protein features on a multi-layer network with 107 tissues. \ohmnet models tissue interdependence up and down a tissue hierarchy spanning dozens of biological scales. The learned features achieve excellent accuracy on the cellular function prediction task, allow us to transfer functions to unannotated tissues, and provide insights into tissues. 

There are several directions for future work. Our approach assumes the dependencies between layers are given in the form of a hierarchy. In several biological scenarios, the dependencies are given in the form of a graph, and we hope to extend the approach to handle graph-based dependencies. As the learned protein features are independent of any downstream task, it would be interesting to see whether our approach performs equally well for gene-disease association prediction and disease pathway detection.

\section*{Funding}

This research has been supported in part by NSF IIS-1149837, NIH BD2K U54EB020405, DARPA SIMPLEX N66001 and Chan Zuckerberg Biohub.

\vspace{3mm}

{\em \noindent Conflict of Interest:} none declared.

\bibliographystyle{natbib}
\bibliography{zitnik.bib}

\end{document}